\documentclass{article}

\usepackage{tccml_iclr2025_conference,times}

\usepackage[inline]{enumitem}
\usepackage{booktabs}
\usepackage{multirow}
\usepackage{graphicx}
\usepackage{subfigure}
\usepackage{arydshln}
\usepackage{wrapfig}
\usepackage{siunitx}

\usepackage{amsthm,amsmath,amssymb,mathrsfs}
\usepackage[colorlinks=true,linkcolor=blue,urlcolor=blue,citecolor=blue]{hyperref}

\usepackage{pifont}

\graphicspath{plots/}

\title{A synthetic dataset of French electric load curves with temperature conditioning}

\author{Tahar Nabil\textsuperscript{1}, Ghislain Agoua\textsuperscript{1},
Pierre Cauchois\textsuperscript{2}, Anne De Moliner\textsuperscript{2}, Beno{\^i}t Grossin\textsuperscript{1}\\
\textsuperscript{1} EDF R\&D, Palaiseau, France\\
\textsuperscript{2} Enedis, France\\
\textit{\{tahar.nabil, ghislain.agoua, benoit.grossin\}@edf.fr}}

\iclrfinalcopy

\begin{document}

\maketitle{}

\begin{abstract}
The undergoing energy transition is causing behavioral changes in electricity use, e.g. with self-consumption of local generation, or flexibility services for demand control.
To better understand these changes and the challenges they induce, accessing individual smart meter data is crucial.
Yet this is personal data under the European GDPR.
A widespread use of such data requires thus to create synthetic realistic and privacy-preserving samples.
This paper introduces a new synthetic load curve dataset generated by conditional latent diffusion.
We also provide the contracted power, time-of-use plan and local temperature used for generation.
Fidelity, utility and privacy of the dataset are thoroughly evaluated, demonstrating its good quality and thereby supporting its interest for energy modeling applications.
\end{abstract}

\section{Introduction} \label{sec:intro}

Meeting the Net Zero Emissions by 2050 Scenario calls for a swift energy transition, involving more renewables on the production side and an intense electrification on the demand side \citep{iea2024gecm}. Optimizing supply and balance, developing new energy services; e.g. around self-consumption of local renewable production or demand control, become thus both increasingly challenging and dependent on one key resource: access to fine-grained electricity consumption data at the individual level \citep{chen2023sm}.
However, strict privacy regulations severely limit the sharing of smart meter data \citep{enedis2024privacy}.
This creates a significant barrier for the research community to develop innovative models for load curve analysis.

On the other hand, the remarkable successes of deep learning generative models in several domains has contributed to initiate a promising area of research:
learning to generate synthetic yet realistic smart meter data, while ensuring their privacy \citep{chai2024defining}.
Nevertheless, the available synthetic datasets suffer from certain shortcomings.
Specifically, they are either 
\begin{enumerate*}[label=(\roman*)]
\item coarse-grained at the hourly timestep and building level \citep{emami2023buildingsbench}, lacking the richness observed at the individual sub-hourly levels; or
\item restricted to daily profile data \citep{liang2022synthesis,yuan_synthetic_2023,thorve_high_2023,chai2024faraday} with no conditioning on temperature, whereas the latter has a significant impact on demand in countries where electric space heating (e.g. France) or cooling prevails.
\end{enumerate*}

In this work, we release a new synthetic dataset of one-year 30-min resolution electric load curves by leveraging latent diffusion models \citep{rombach2022high}.
We extend their conditioning mechanism to time-varying exogenous variables.
This allows us to condition our samples on outdoor temperature, also released.
A thorough evaluation shows that the proposed model
\begin{enumerate*}[label=(\arabic*)]
\item yields unprecedented high-quality samples, with superior performance compared to TimeGAN, a standard time series generative model, without compromising privacy;
\item can flexibly incorporate any static information describing the consumer, enabling us to feature a wide range of customer contracted powers and individual behaviors.
\end{enumerate*}
\section{Conditional load curve generation with Latent Diffusion} \label{sec:model}

Our dataset is generated by training a conditional latent diffusion model \citep{rombach2022high}, which we describe next.
A brief discussion of related public smart meter datasets is provided in Appendix \ref{app:gen-models}.
Our architecture is depicted in Supplementary Figure \ref{fig:archi}.

\paragraph{Latent diffusion model (LDM)}
Load curves $\mathbf{x}\in\mathbb{R}^{T\times1}$, with $T=N_{days}\times48$ at sampling rate 30 minutes, are seen as single-channel 2D images ${\tilde{\mathbf{x}}}\in\mathbb{R}^{1\times N_{days} \times 48}$.
LDMs are two-stage models.
(1) First, we fit a convolutional autoencoder with compression factor $r=4$ to reconstruct $\tilde{\mathbf{x}}$.
$\mathbf{x}$ is therefore represented by a latent code $\mathbf{z}\in\mathbb{R}^{c_z\times N_{days}/r \times 48/r}$ of $c_z$ channels.
The loss function of the autoencoder adds a vector quantization term to the L2 reconstruction loss to regularize the latent space \citep{rombach2022high,vandenoord2017neural}.
(2) Next, the distribution $p(\mathbf{z})$ of codes in the latent space of the frozen autoencoder is approximated by learning a Denoising Diffusion Probabilistic Model (DDPM, \cite{ho2020denoising}).
Specifically, a UNet with spatial attention is trained to minimize the denoising objective.

\paragraph{Conditioning mechanisms}
Leveraging the flexibility of latent diffusion, we propose to condition the model on both static variables and other time-varying exogenous variables.
\begin{enumerate}
\item \emph{Static variables:}
following \cite{rombach2022high}, the UNet model is conditioned on labels by either concatenation to $\mathbf{z}$ or cross-attention through a spatial transformer.
\item \emph{Exogenous variables:}
to condition $\mathbf{x}$ on another series $\mathbf{u}\in\mathbb{R}^{T\times 1}$,
we process its code $\mathbf{z}$ with a cross-attention layer, with $\mathbf{z}$ as the query and patched $\mathbf{u}\in\mathbb{R}^{(T//P)\times P}$ as the keys and values, where  $P$ is the patch length \citep{nie2023time}.
This results in a transformed code $\mathbf{h}\in\mathbb{R}^{c_z\times N_{days}/r \times 48/r}$ which is then processed by the decoder.
\end{enumerate}
Hence \begin{enumerate*}[label=(\roman*)]
\item conditioning on exogenous variables such as outdoor temperature is handled by the decoder only and does not affect the diffusion loss for learning the distribution of load curves; but
\item conditioning on static variables does not affect fitting the autoencoder for obtaining good reconstructions.
Provided that the autoencoder generalizes correctly, any static conditioning can thus be added in the second stage of learning the data distribution.
\end{enumerate*}

\paragraph{Generation}
At inference, the diffusion process is reversed to sample a code $\hat{\mathbf{z}}$ conditionally on labels.
$\hat{\mathbf{z}}$ undergoes cross-attention with the exogenous variables and is then decoded to data space.

\section{Evaluation} \label{sec:eval}

We compare our latent diffusion model (LDM) to TimeGAN, the standard baseline for time series generation \citep{yoon2019time}.
Both are trained on a set of 17k one-year French load curves, conditionally on temperature and on two labels of 3 classes each:
contracted power and time-of-use (ToU) rate.
They are evaluated along the three key dimensions of synthetic data: fidelity (Section \ref{ssec:fidelity}), utility (in \ref{ssec:utility}) and privacy (in \ref{ssec:privacy}).
Full data description, models implementation, metrics and results are given respectively in Appendices \ref{app:dataset}, \ref{app:model}, \ref{app:metrics} and \ref{app:results-full}.

\subsection{Fidelity \& Diversity} \label{ssec:fidelity}

The evaluation relies on three metrics.
\begin{enumerate*}[label=(\roman*)]
\item The discriminative score \citep{yoon2019time} is the test accuracy of a \emph{post-hoc} 1-NN binary classifier separating real and synthetic data with Euclidean distance.
\item Context-FID measures how well the synthetic dataset recovers the training set statistics in a suitable embedding space \citep{jeha2022psagan, franceschi2019unsupervised}.
\item The correlation score measures temporal consistency \citep{ni2022sig}.
\end{enumerate*}

\begin{table}[]
\centering
\caption{Fidelity scores on the hold-out test set for samples on \emph{night} ToU and 6 kVA.
Discriminative score computed on one-year load curves ($D_{year}$) or on averaged daily profiles ($D_{profile}$).
Best performance emphasized in bold.
Full results in Supplementary Table \ref{tab:fidelity-all}.
}
\begin{tabular}{ccccc}
\toprule
& $D_{year}$ ($\downarrow$) & $D_{profile}$ ($\downarrow$) & \bfseries Context-FID ($\downarrow$) & \bfseries Correlation score ($\downarrow$)\\
\midrule
\bfseries LDM     & \bfseries 0.037 & \bfseries 0.059 & \bfseries 1.748 & \bfseries 0.002 \\
\bfseries TimeGAN & 0.357 & 0.452 & 2.082
& 0.224 \\
\bottomrule
\end{tabular}
\label{tab:fidelity}
\end{table}

\begin{figure}
\centering
\subfigure[t-SNE]{
\raisebox{3mm}{\includegraphics[width=0.205\textwidth]{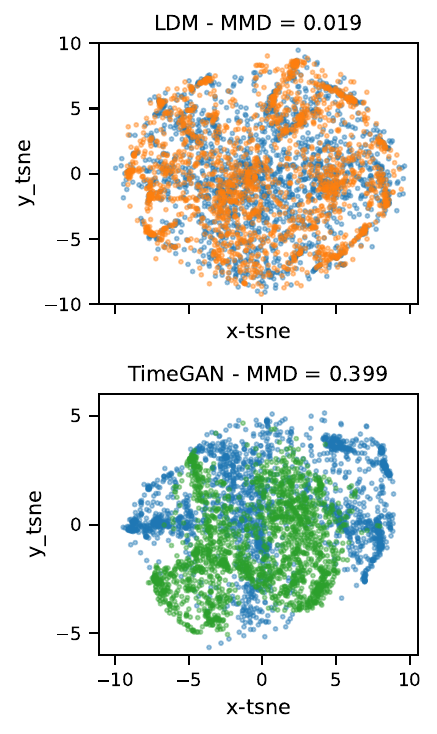}}
}
\subfigure[Statistics and Quantiles]{
\includegraphics[width=0.75\textwidth]{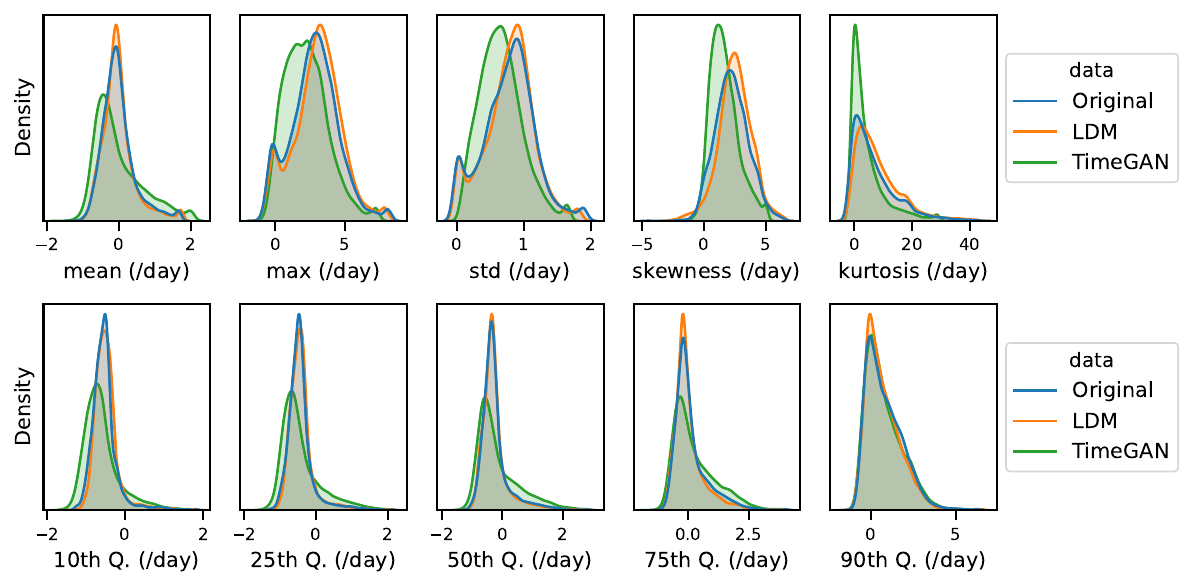}
}
\\
\vspace{-2ex}
\subfigure[Individual sample (30-day zoom)]
{
\raisebox{-1mm}{\includegraphics[width=0.475\textwidth]{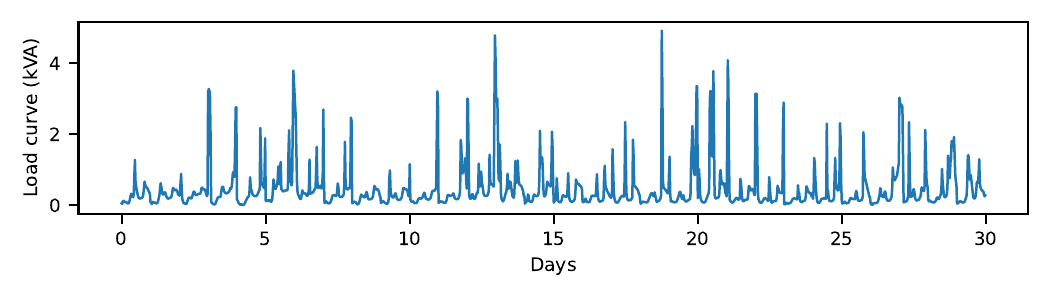}}
}
\subfigure[Weekly profiles]
{
\includegraphics[width=0.475\textwidth]{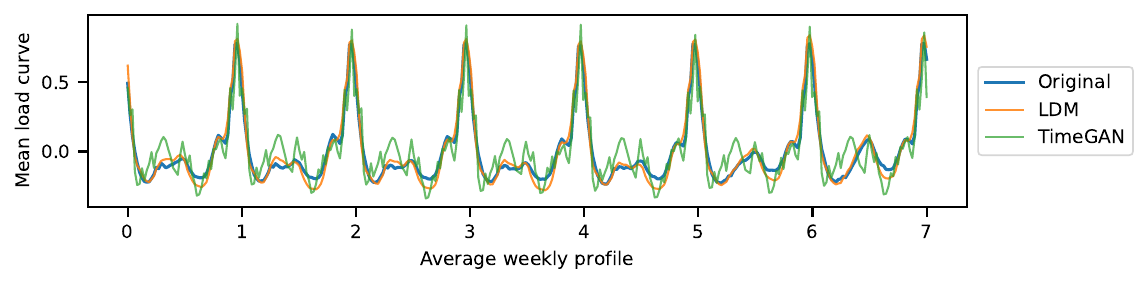}
}
\caption{
(a)
t-SNE 2D projection of original (blue) and synthetic data across all categories for Latent Diffusion (orange) and TimeGAN (green).
(b) to (d) are restricted to the night ToU, 6kVA category:
(b) Density estimation of daily statistics and quantiles;
(c) Example of a synthetic sample by Latent Diffusion;
(d) Mean weekly profiles.
}
\label{fig:plots}
\end{figure}

\paragraph{Results}
The metrics shown in Table \ref{tab:fidelity} and Supplementary Table \ref{tab:fidelity-all}, demonstrate that LDM consistently outperforms TimeGAN, with high-quality samples challenging to distinguish from real data.
Qualitatively, \begin{enumerate*}[label=(\roman*)]
\item the 2D-projection of the samples with t-SNE in Figure \ref{fig:plots}(a) is consistent with the discriminative scores,
with LDM samples hard to distinguish from real samples;
\item LDM outperforms TimeGAN as shown by the histograms of daily statistics of interest (mean, max, quantiles, etc.) in Figure \ref{fig:plots}(b), by the average weekly profiles in Figure \ref{fig:plots}(d), as well as by additional plots in Appendix \ref{app:supp-fidelity-plots}.
\item Some LDM individual samples are displayed in Figures \ref{fig:plots}(c) and \ref{fig:samples6}-\ref{fig:samples12}, showcasing its ability to generate diverse patterns, e.g. plateaus of low consumption, while respecting known seasonalities (daily peaks).
\end{enumerate*}

\paragraph{Thermo-sensitivity}
We further evaluate whether the synthetic data preserve the correlation between temperature and consumption.
Quantitatively, we compare the distributions of the thermo-sensitivity gradients, which explain daily variations in consumption in winter by those in temperature (see Appendix \ref{gradient}).
Qualitatively, we generate load curves under a strong global temperature offset throughout the year.
The results, in Appendix \ref{app:results-thermo}, suggest that the latent diffusion model learned a meaningful conditioning by temperature, with (i) gradients matching the real test distribution and (ii) realistic distortion of the load curves under the temperature offset.
\subsection{Utility} \label{ssec:utility}

Utility assesses whether generated data can be used to solve machine learning tasks to the same extent as with real data.
Beyond next-step prediction \citep{yoon2019time}, new tasks must be designed for electricity consumption \citep{chai2024defining}.
We adopt the standard \emph{Train on Synthetic, Test on Real} (TSTR) setting, where \emph{post-hoc}
models are trained either on synthetic data or real train data, and compared on a hold-out real test set.
We tackle two tasks, short-term load forecasting and time series classification.

\paragraph{Forecasting}
Using a lookback-window of 15 days, we train a \emph{post-hoc} state-of-the-art transformer model, PatchTST \citep{nie2023time}, to predict individual load curves for various horizons $H$, from a day ($H=48$) to a week ($H=336$).
Table \ref{tab:utility} shows the mean squared (MSE) and absolute (MAE) errors, averaged across all horizons (detailed scores in Supplementary Table \ref{tab:forecast-full}):
models trained on either real or latent diffusion data are in close agreement, whereas training on TimeGAN degrades the MSE on real test data by about 9\%.

\paragraph{Classification}
We train a kNN (k=5) classifier to predict the ToU (3 classes) associated to each load curve.
The input representation of the classifier is a 104-dimensional vector, where each load curve is transformed into a vector of daily profile over the year (48 values), daily profile restricted to winter (48 values) and a vector of 8 statistics.
The ground truth for fake data is the conditioning label.
The test accuracy and F1-scores in Table \ref{tab:utility} yield conclusions similar to the forecasting task, confirming that LDM produces high-quality data.

\begin{table}
\centering
\caption{TSTR metrics; the closest model to TRTR (\emph{Train on Real, Test on Real}) is emphasized in bold.
Forecasting results are averaged across horizons $[48,96,192,336]$, for a lookback of length 720.
Baselines: copy from last week (forecasting), majority class (classification).
}
\begin{tabular}{r|c|cc|c}
\toprule
\bfseries Loss& \bfseries TRTR & \bfseries LDM & \bfseries TimeGAN & \bfseries Baseline \\
\midrule
\bfseries\small Forecast. MSE & 0.190 & \bfseries 0.190 & 0.209 & 0.306 \\
\bfseries\small Forecast. MAE & 0.234 & \bfseries 0.233 & 0.253 & 0.251 \\
\midrule
\bfseries\small Classif. Acc & 0.740 & \bfseries 0.750 & 0.700 & 0.607 \\
\bfseries\small Classif. F1  & 0.576 & \bfseries 0.564 & 0.537 & 0.252 \\
\bottomrule
\end{tabular}
\label{tab:utility}
\end{table}

\subsection{Privacy} \label{ssec:privacy}

We evaluate the privacy of synthetic samples through two Membership Inference Attacks (MIA, \cite{carlini2022membership}) and a statistical test of relative similarity (MMD-test).

\paragraph{MIA}
Commonly considered as the first step towards auditing the privacy of deployed models,
MIAs try to predict whether a given sample has been seen by the target model during its training.
We perform two attacks, described in Appendix \ref{app:metrics}:
\begin{enumerate*}[label=(\roman*)]
\item black-box, by exploiting the distance to generated samples; 
\item white-box, by exploiting the reconstruction scores of the autoencoders.
\end{enumerate*}
Following best practices \citep{carlini2022membership}, we report the ROC curve in log-scale in Supplementary Figure \ref{fig:mia-roc} and the true positive rate at 0.1\% false positive rate in Supplementary Table \ref{tab:mia-privacy}.
The performance of the attacks are close to random for both models, at every false positive rate.

\paragraph{MMD-test}
In addition, we perform a three-sample MMD test with the following null hypothesis:
the distribution of synthetic daily profiles is closer to test than train data \citep{bounliphone2016test}.
The p-values in Supplementary Table \ref{tab:mmd-privacy} are large ($>0.2$ for all categories but one at 0.019),
suggesting no evidence for the models to overfit training data.
Similar conclusions apply to TimeGAN.

Together with the computation of the nearest neighbor distance ratio (Appendix \ref{app:res-privacy}), these tests converge thus to the same conclusion:
without being a formal proof of privacy, they are hints that the model is not prone to overfitting and yields original synthetic samples, not copying the train set.
\section{Conclusion and Impact Statement} \label{sec:conclusion}
This work proposes a new and richer synthetic dataset for residential load curves, covering one year, with contracted power and ToU plan information, as well as the local temperature data used for generation. The generation is achieved by leveraging the flexibility of latent diffusion models and an astute conditioning mechanism that incorporates both static and dynamic conditional variables. The evaluation shows good fidelity, utility and privacy metrics. 
Most notably, LDM consistently achieves high-quality scores in every fidelity aspect.

Future work could extend the generation to more categories, e.g. non-thermo-sensitive load curves or small and medium-sized businesses.
Besides, properly assessing the utility of smart meter data remains an open question:
other ideas could exploit the conditioning on temperature to forecast with exogenous variables \citep{wang2024timexer}, perform transfer experiments \citep{emami2023buildingsbench} or train time series foundation models \citep{woo2024unified}.

\subsection*{Data Availability}
We publicly release a synthetic dataset (\href{http://dx.doi.org/10.5281/zenodo.15232742}{doi: 10.5281/zenodo.15232742}) of 10k one-year residential load curves, with their local temperatures, contracted power and ToU plans.
This open dataset generated by conditional latent diffusion captures the correlation between cold temperature and electric consumption that can be found in certain countries, e.g. in France.

\subsection*{Acknowledgement}
This paper is based on joint work between Enedis and EDF R\&D in accordance with regulated R\&D contract.
Moreover, we would like to thank Etienne Le Naour for the valuable discussions on this project and feedbacks on the paper.

\bibliography{main}
\bibliographystyle{abbrvnat}

\newpage
\appendix
\section{Background} \label{app:background}

\subsection{Related work}  \label{app:gen-models}

\paragraph{Open electricity consumption data}
First open electricity consumption (load curve) datasets typically contained real smart meter readings for the aggregated household consumption at sub-hourly timesteps, e.g. 15 minutes (Electricity, in Portugal, \cite{electricity2015}) or 30 minutes (\href{https://www.ucd.ie/issda/data/commissionforenergyregulationcer}{Irish Commission for Energy Regulation dataset}, \href{https://data.london.gov.uk/dataset/smartmeter-energy-use-data-in-london-households}{UK Power Networks (London)}, \href{https://data.gov.au/dataset/ds-dga-4e21dea3-9b87-4610-94c7-15a8a77907ef/details}{Smart-Grid Smart-City Customer Trial Data, Australia}).
However, they are of limited use to grid operators and others involved in the energy transition, dating back to around a decade ago and corresponding thus to other user behaviors.
Similarly, \href{https://data.enedis.fr/explore/dataset/conso-inf36/information/?disjunctive.profil&disjunctive.plage_de_puissance_souscrite}{{conso-inf36}} is an open dataset of regularly updated 30-min resolution French load curves, but limited to averages of at least 100 individuals.
Recent works have investigated the use of generative models to produce synthetic realistic consumption data, e.g. \citep{liang2022synthesis,yuan_synthetic_2023,thorve_high_2023,chai2024faraday}.
These references restrict themselves to daily profiles at hourly or sub-hourly timesteps.
For instance, Faraday generates daily profiles with a Variational Auto-Encoder and provides information such as property type and low carbon technologies ownership (e.g. electric vehicles ownership) \citep{chai2024faraday}.
A common limitation of all aforementioned references is the lack of conditioning by outdoor temperature.
Finally, another line of work relies on simulation:
BuildingsBench \citep{emami2023buildingsbench} simulates a large number of one-year hourly residential and commercial building load curves, representative of the US stock, conditionally on local temperature -- without releasing the temperature time series.

\subsection{Data details} \label{app:dataset}
\paragraph{Training data}
The available training dataset contains the following variables:
\begin{itemize}
\item 17k residential load curves spanning 94 departments in metropolitan France.
The individuals have time-of-use (ToU) plan, their consumption is correlated with outdoor temperature, particularly during winter. Data is at 30-min resolution and covers one year starting from October 2022;
\item Local outdoor temperature data for the same period;
\item Contracted power information: 6, 9, 12 (kVA);
\item Time-of-use (ToU) plan: 3 classes "midday", "night" and "misc" describe the time of day when prices are lower.
ToU strongly conditions the load curve, in particular the daily patterns and the time of peak consumption.
\end{itemize}
Test data contain 2k samples with the same information as for training.

\subsection{Thermo-sensitivity gradients} \label{gradient}

When aggregated at a daily granularity, the analysis of load data and temperature shows an almost linear relationship between load and temperature below a temperature threshold.  The thermo-sensitivity gradient represents the relationship between load and temperature below the threshold. The steps to calculate the gradient for a daily load curve are as follows: 
\begin{enumerate}
\item filter on winter data (from November to March) to avoid bias;
\item calculate the load delta: $\Delta Load_d = Load_d - Load_{d-7}$ where $Load_d$ is the load for day $d$ in kWh;
\item calculate the degree-day $ DJU_d = max(0, T_{thresh} - T_d)$ where $T_d$ is the temperature for day $d$ and $T_{thresh} \in [14.5, 18]\,(\text{in }\SI{}{\celsius})$ the temperature threshold for the region;
\item calculate the delta of degree-days $\Delta DJU_d = DJU_d - DJU_{d-7}$;
\item compute the linear regression $\Delta Load = Gradient \times \Delta DJU$.
\end{enumerate}
The gradients are in kWh/degree-day.

\section{Model and baseline details} \label{app:model}

\subsection{Latent Diffusion} \label{app:ldm}

\begin{figure}
\centering
\includegraphics[width=0.85\textwidth]{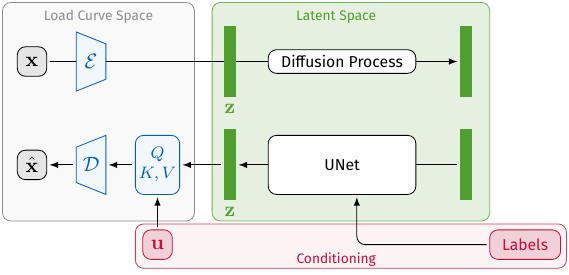}
\caption{Latent diffusion \citep{rombach2022high} with conditioning on exogenous variables.
$\mathcal{E}$: image encoder,
$\mathcal{D}$: image decoder,
$Q,K,V$: cross-attention network between latent vector $\mathbf{z}$ and exogenous time series $\mathbf{u}$.
$\mathbf{x}$ is the input data (load curve) and $\hat{\mathbf{x}}$ its reconstruction.
Static labels are added by cross-attention to the diffusion UNet, or by concatenation to the latent codes $\mathbf{z}$.
}
\label{fig:archi}
\end{figure}

\paragraph{Architecture and hyperparameters}
The overall architecture is represented in Figure \ref{fig:archi}.
The hyperparameters of the first stage autoencoder model are:
\begin{itemize}
\item Compression ratio: $r=4$;
\item Input channels: 1;
\item Number of channels: 128, channel multipliers: $[1, 2, 4]$;
\item Codebook: number of channels $c_z=3$, shape $3\times90\times12$, cardinality: 8192;
\item Temperature conditioning:
\begin{itemize}
\item Temperature (in \SI{}{\celsius}) is scaled by \SI{35}{\celsius} and reshaped as non-overlapping patches of size 32;
\item Cross-attention network:
attention heads: 8,
attention layers: 4,
positional encoding with sines and cosines;
\end{itemize}
\item  Model size: 61M parameters.
\end{itemize}

The hyperparameters for the denoising diffusion model are:
\begin{itemize}
\item Diffusion steps: 1000;
\item Noise schedule: linear;
\item UNet architecture (spatial attention):
\begin{itemize}
\item Attention heads: 16;
\item Number of channels: 160, multipliers: $[1,2,3,4]$;
\end{itemize}
\item Label representation:
contracted power normalized between 0 and 1 is concatenated to a learned embedding of dimension 4 of the tariff (3 classes);
\item Label conditioning: concatenation to the latent code along the channel dimension;
\item Model size: 157M parameters.
\end{itemize}

\paragraph{Implementation details}
Load curves are z-normalized instance-wise, by computing the mean and standard deviation across the time dimension, for each sample.
The autoencoder is trained with Huber loss, a learning rate of $5\times10^{-5}$ for 250k optimization steps of batch size 16 and a cosine annealing schedule with 10k warmup steps.
We also add data augmentation when training the autoencoder:
with probability 0.5, we shift the temperature by a random global offset $\delta$, and modify consequently the load curve by adding $-g\times\delta$, with $g>0$ a random thermo-sensitivity gradient.
The diffusion model is trained with mean squared error loss for the denoising objective (see \cite{ho2020denoising,rombach2022high}), for 200k steps with batch size 16, learning rate $3\times10^{-6}$, cosine annealing with 10k warmup steps.
For both the autoencoder and the diffusion model, we used the AdamW algorithm to optimize the loss functions \citep{loshchilov2018decoupled}, with default parameters $\beta_1=0.9,\,\beta_2=0.999,\,\epsilon=10^{-8},\,\lambda=0.01$.
All models were trained on a single NVIDIA A100-40GB GPU.

\subsection{TimeGAN}
The TimeGAN model \citep{yoon2019time} is a Generative Adversarial Network model that operates in the latent space of an autoencoder. Therefore, it consists of 4 networks: encoder, decoder, generator and discriminator, which are all GRUs \citep{cho2014gru} to better handle the temporal dynamics. 
Similarly to Latent Diffusion, each time series $\mathbf{x}\in\mathbb{R}^{T\times1}$ is represented as a sequence of $N_{days}$ 48-dimensional vectors $\tilde{\mathbf{x}}\in\mathbb{R}^{N_{days}\times 48}$, with $T=N_{days}\times48$.
Conditioning by temperature is achieved by concatenation of the temperature to the inputs of all 4 sub-networks of TimeGAN, along the time dimension.
Similarly for conditioning by discrete labels, which are repeated $N_{days}$ times and stacked along the second time dimension as well.
The input load curve data is transformed through an inverse hyperbolic sine (\emph{asinh}) transformation.
Load curve and temperature are then normalized with a min-max scaler.
The parameters of the models were optimized using Optuna \citep{optuna2019} and grid search. The final parameters are:
\begin{itemize}
\item hidden dimension $h_{dim}=32$ for the latent space of the autoencoder, $z_{dim}=24$ for the generator and $n_{layers}=1$ layer in all GRU's networks;
\item we use different learning rates $lr=0.0005$ for the encoder, $lr_{gen}=0.0005$ for the generator, and $lr_{disc}=0.005$ for the discriminator to increase the stability of the training;
\item the model is trained with a batch size of $32$ and $15$k steps.
\end{itemize}

\section{Evaluation protocol} \label{app:metrics}

\paragraph{Description}
The evaluation of the generative models relies on a standard protocol, by comparing sets of synthetic data to a hold-out test set of actual load curves.
We generate as many samples per conditioning label as there is in the test set, and use the temperatures in the test set:
both synthetic and test sets have the same number of samples $n_{test}$.
The two sets are compared along the three dimensions:
fidelity, utility and privacy.

\paragraph{Fidelity metrics}
Fidelity metrics assess whether the synthetic data are realistic and diverse.
The following metrics are computed:
\begin{itemize}
\item \emph{Discriminative score:}
following \cite{yoon2019time}, we use $n_{test}/2$ original and $n_{test}/2$ synthetic samples to train a binary classifier, where original (respectively, synthetic) data are labelled 0 (resp., 1).
We compute the accuracy $\mathrm{acc}$ on the rest of the original and generated data and report $|1/2-\mathrm{acc}|$ as the discriminative score.
Perfectly separable original and synthetic sets are expected to achieve a score of 0.5; whereas indistinguishable sets are close to 0.
We use a 1-nearest-neighbor with Euclidean distance as classifier, computed either in the space of yearly data ($D_{year}$) or after averaging the load curves to daily profiles of 48 values ($D_{profile})$.
\item \emph{Context-FID} is a Frechet-Inception distance-like score for time series.
In computer vision, FID assseses the quality of images by comparing the distribution of synthetic vs real images, through the means and covariance matrices of their respective embeddings, produced by an Inception network.
Here, following \cite{jeha2022psagan},
the embedding network consists in temporal convolutions with triplet loss (contrastive learning with an encoder-only architecture, from \cite{franceschi2019unsupervised}).
We train one network per dataset, synthetic or real.
To better embed local contexts, we split the one-year series into one-month series and fit one model per month.
We report the average of the FID scores computed for all 12 months.
Best scores are the lowest.
\begin{itemize}
\item The hyperparameters for training the encoder network are:
10k steps with batch size 128, learning rate = $10^{-4}$, 10 layers of dilated causal 1D convolutions with 40 channels, output convolution with 320 channels, linear projection to a latent dimension of 160.
The final max-pooling layer across sequence dimension guarantees that all series are encoded as vectors in $\mathbb{R}^{160}$.
\end{itemize}
\item \emph{Correlation score:}
assesses temporal consistency by computing the absolute difference between auto-correlation functions of real and synthetic data
\citep{ni2022sig}.
\end{itemize}

\paragraph{Utility metrics}
Utility metrics measure whether synthetic data are useful for solving other downstream machine learning tasks.
They are computed in the \emph{TSTR, Train on Synthetic, Test on Real} scenario, i.e by training a model on synthetic data and testing it on a hold-out test set of real data.
The test metrics are compared to the same model trained on real data (\emph{TRTR, Train on Real, Test on Real}).
\begin{itemize}
\item Forecasting task: the hyperparameters of the forecasting model, PatchTST \citep{nie2023time} are as follows:
\begin{itemize}
\item Lookback window: 720, horizon $H\in[48,96,192,336]$
\item Patch length: 32, stride: 16
\item Attention: 3 layers, 16 heads and $d_{model}=128$ (token dimension)
\item Model trained with reversible instance normalization (RevIN, \cite{kim2022reversible})
\item Learning rate: $10^{-6}$
\item Batch size: 256, number of steps: 20k, cosine learning rate scheduler with 1k warmup steps
\end{itemize}
\end{itemize}
For fairness, all training sets in TSTR or TRTR are restricted to the same size.
TRTR results are averaged over 5 runs from random subsamples of the training set.

\paragraph{Privacy metrics}
\begin{itemize}
\item \emph{Black-box membership inference attack:}
Membership inference attacks are fundamental privacy attacks of machine learning models, trying to predict if a particular sample has been seen during training.
They are the foundations for stronger attacks \citep{carlini2022membership}.
In a black box scenario, the attacker has only access to queries to the target model under attack, and is given a set $\mathbf{x}_1,\dots\mathbf{x}_{n+m}$ of $n+m$ samples of which $n$ are known to be from the training set.
We perform a simple attack:
the attacker decides to classify as train the samples with score below a threshold, and as test otherwise.
In this blackbox scenario, the score is the minimum distance between the target sample and any synthetic sample.
\item \emph{White-box (loss) membership inference attack:}
assumes that the attacker has access to the trained machine learning model - although in our case, the model is not publicly available.
The attack is similar to blackbox, except that the score is the reconstruction error of the samples by the autoencoder \citep{hayes2017logan}.
\item \emph{Three-sample Maximum Mean Discrepancy Test:}
initially, \cite{bounliphone2016test} introduced a statistical test of relative similarity to determine which of two models generates samples that are significantly closer to a real-world reference dataset of interest.
The test statistics are based on differences in maximum mean discrepancies (MMDs).
Here, we use  the test to assess whether the train set is significantly closer to the generated samples than the test set.
In other words, our three samples are:
the synthetic dataset and the original train and test sets.
We perform the test in the space of daily profiles instead of one-year curves because of the cost of MMD computation and since profiles are easier to discriminate.
The null hypothesis is thus $\mathcal{H}_0:$
the distribution of synthetic daily profiles is closer to test daily profiles than train daily profiles.
See \citep{bounliphone2016test} for more details on the implementation of the test.
Rejecting $\mathcal{H}_0$ would imply that there is statistical evidence that the generative model has overfitted the train set, which would be a breach of privacy.
\item \emph{Nearest Neighbor Distance Ratio (NNDR):}
given a real train (respectively, test) sample, we compute the NNDR as the ratio of (i) the distance to its nearest neighbor in the synthetic set to (ii) the distance to its second nearest neighbor, in the synthetic set.
Values of NNDR close to 1 suggest that synthetic samples are located in dense regions of real data, whereas values close to 0 indicate that some synthetic samples are located in sparse regions, i.e. near outliers.
\end{itemize}

\section{Full experimental results} \label{app:results-full}

\subsection{Fidelity}

\subsubsection{Metrics and plots} \label{app:supp-fidelity-plots}

\begin{table}
\centering
\caption{Fidelity scores on the hold-out test set for all categories of contracted power and time-of-use (ToU).
Discriminative score computed on the entire load curves ($D_{year}$) or on averaged daily profiles ($D_{profile}$).
Best performance emphasized in bold.}
\begin{tabular}{cc|c|cccc}
\toprule
\bfseries Contracted& \multirow{2}{*}{\bfseries ToU} & \multirow{2}{*}{\bfseries Model} & $D_{year}$ & $D_{profile}$ & \bfseries Context-& \bfseries Correlation\\
\bfseries Power&&& ($\downarrow$) & ($\downarrow$) & \bfseries FID ($\downarrow$)  & \bfseries score ($\downarrow$) \\
\midrule
\multirow{6}{*}{6 kVA} & \multirow{2}{*}{midday} & LDM & 
\bfseries 0.049 & \bfseries 0.078 & \bfseries 2.394 & \bfseries 0.003
\\
& & TimeGAN & 0.388 & 0.471 & 2.802 & 0.130 \\
\cdashline{2-7}
& \multirow{2}{*}{night} & LDM & \bfseries 0.037 & \bfseries 0.059 & \bfseries 1.748 & \bfseries 0.002 \\
& & TimeGAN & 0.357 & 0.452 & 2.082 & 0.224 \\
\cdashline{2-7}
& \multirow{2}{*}{misc.} & LDM & \bfseries 0.119 & \bfseries 0.0 & \bfseries 3.405 & \bfseries 0.013 \\
& & TimeGAN & 0.357 & 0.405 & 4.021 & 0.113 \\
\midrule

\multirow{6}{*}{9 kVA} & \multirow{2}{*}{midday} & LDM & \bfseries 0.035 & \bfseries 0.054 & \bfseries 2.000 & 0.059 \\
& & TimeGAN & 0.381 & 0.438 & 2.483 & \bfseries 0.021 \\
\cdashline{2-7}
& \multirow{2}{*}{night} & LDM & \bfseries 0.068 & \bfseries 0.051 & \bfseries 1.622 & \bfseries 0.022 \\
& & TimeGAN & 0.414 & 0.434 & 1.968 & 0.127 \\
\cdashline{2-7}
& \multirow{2}{*}{misc.} & LDM & \bfseries 0.030 & \bfseries 0.091 & \bfseries 4.219 & \bfseries 0.022 \\
& & TimeGAN & 0.424 & 0.379 & 4.631 & 0.029 \\
\midrule

\multirow{6}{*}{12 kVA} & \multirow{2}{*}{midday} & LDM & \bfseries 0.024 & \bfseries 0.032 & \bfseries 2.913 & \bfseries 0.011\\
& & TimeGAN & 0.395 & 0.379 & 3.301 & 0.048 \\
\cdashline{2-7}
& \multirow{2}{*}{night} & LDM &  \bfseries 0.071 & \bfseries 0.150 & \bfseries 1.968 & \bfseries 0.012 \\
& & TimeGAN & 0.387 & 0.444 & 2.278 & 0.089 \\
\cdashline{2-7}
& \multirow{2}{*}{misc.} & LDM & \bfseries 0.0 & \bfseries 0.043 & \bfseries 4.744 & \bfseries 0.005 \\
& & TimeGAN & 0.370 & 0.370 & 5.075 & 0.038\\

\midrule
\multirow{2}{*}{all} & \multirow{2}{*}{all} & LDM & \bfseries 0.029 & \bfseries 0.026 & \bfseries 1.235 & \bfseries 0.011 \\
& & TimeGAN & 0.382 & 0.432 & 1.476 & 0.123\\
\bottomrule
\end{tabular}
\label{tab:fidelity-all}
\end{table}

\paragraph{Metrics}
We provide the breakdown of fidelity metrics for all 9 categories of ToU $\times$ contracted power, as well as for the entire synthetic dataset, in Table \ref{tab:fidelity-all}.
Latent Diffusion consistently achieves high-quality discriminative scores in all categories (close to 0 in all categories, and less than $0.07$ in 9/10 cases).
The gap is significant with TimeGAN, which fails to produce samples undistinguishable from the real distribution.
The Context-FID and correlation scores are also strongly in favor of Latent Diffusion over TimeGAN, supporting the former as a better generative model.

\begin{figure}
\centering
\subfigure[6 kVA, night ToU]{\includegraphics[width=0.475\textwidth]{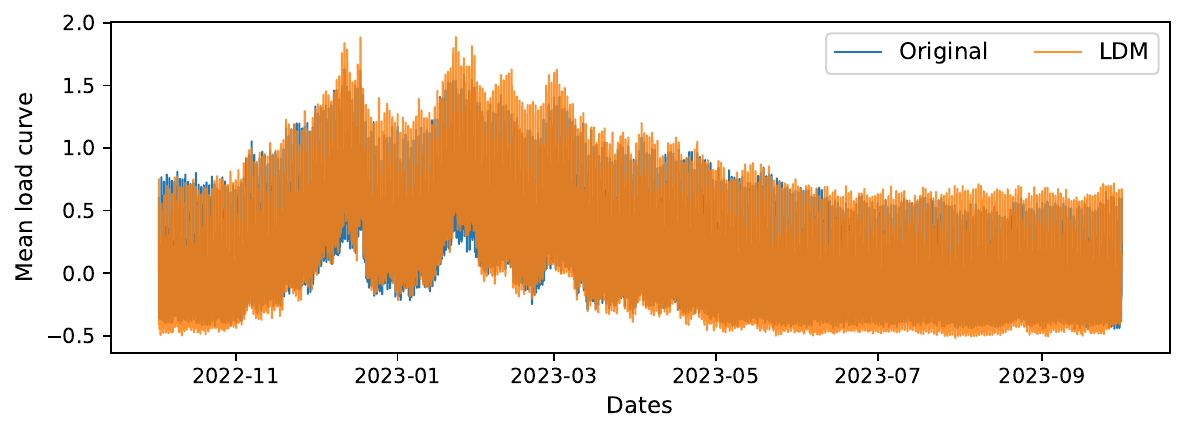}}
\subfigure[6 kVA, night ToU]{\includegraphics[width=0.475\textwidth]{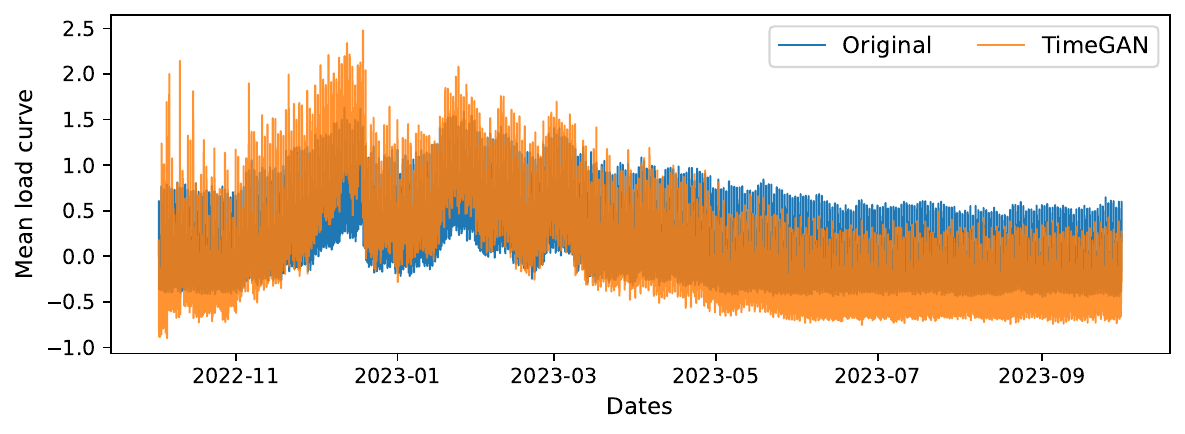}}
\subfigure[All labels]{\includegraphics[width=0.475\textwidth]{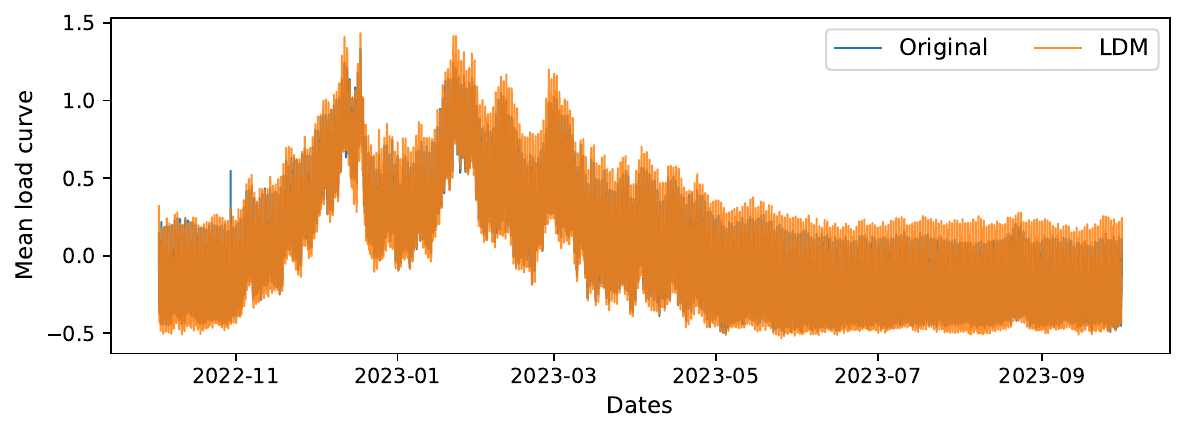}}
\subfigure[All labels]{\includegraphics[width=0.475\textwidth]{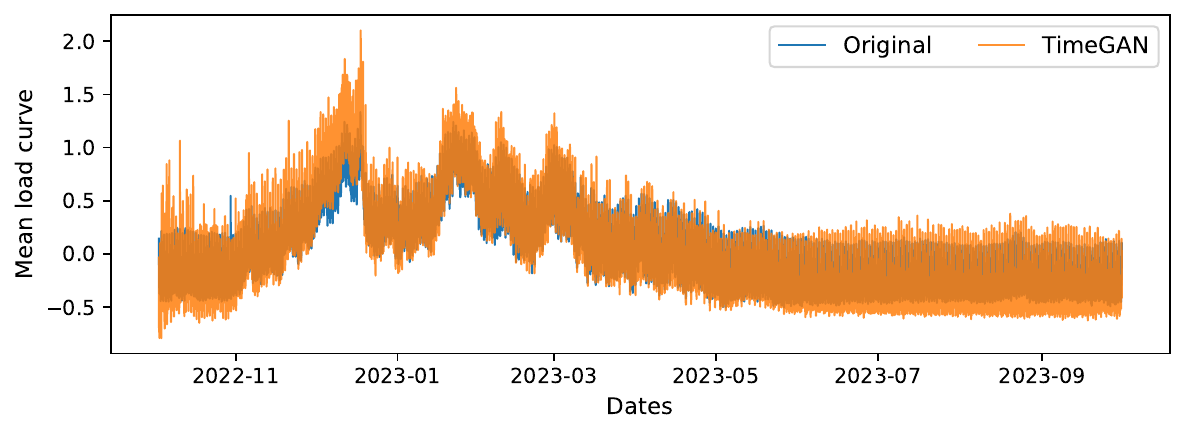}}
\caption{Average one-year load curves of (left) Latent Diffusion and (right) TimeGAN.
}
\label{fig:means}
\end{figure}

\paragraph{Plots}
This section provides additional plots at the aggregated level, i.e. by computing averages over groups of samples.
First, the average one-year load curves is shown in Figure \ref{fig:means}, for one conditioning (6 kVA and night time-of-use) and across all labels.
It can be seen that samples generated by Latent Diffusion are close, in average, to the real test dataset, whereas samples from TimeGAN deviate slightly.
Mean plots of other conditioning categories, not shown here, yield the same conclusions.

\begin{figure}
\centering
\subfigure[6 kVA, night ToU]{\includegraphics[width=0.475\textwidth]{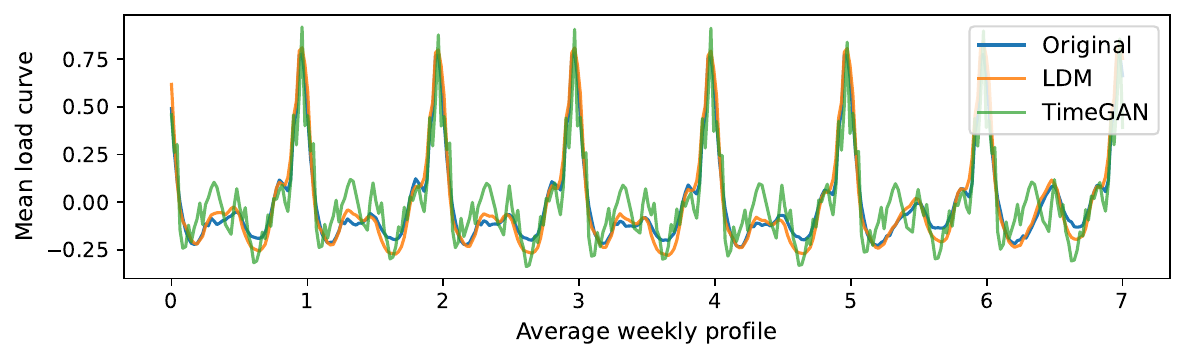}}
\subfigure[9 kVA, midday ToU]{\includegraphics[width=0.475\textwidth]{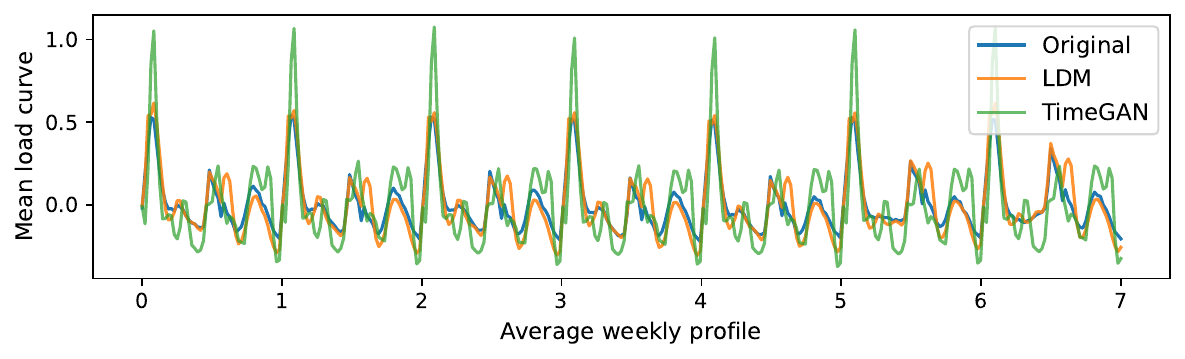}}
\subfigure[12 kVA, misc. ToU]{\includegraphics[width=0.475\textwidth]{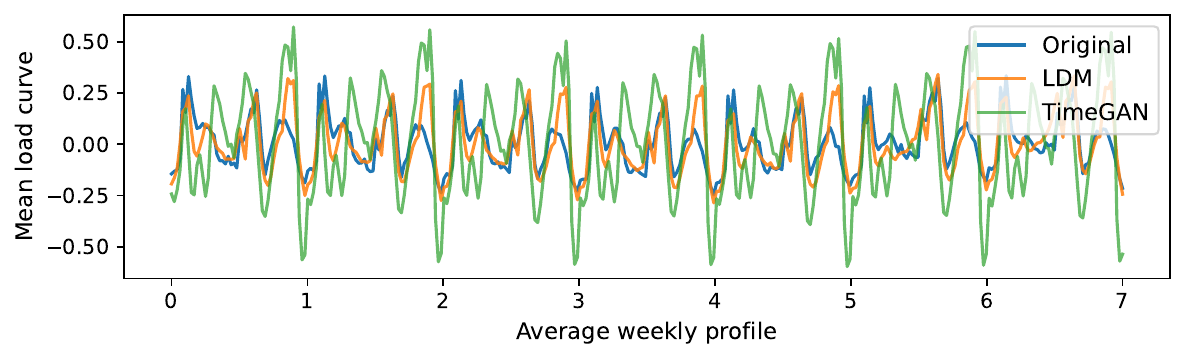}}
\subfigure[All labels]{\includegraphics[width=0.475\textwidth]{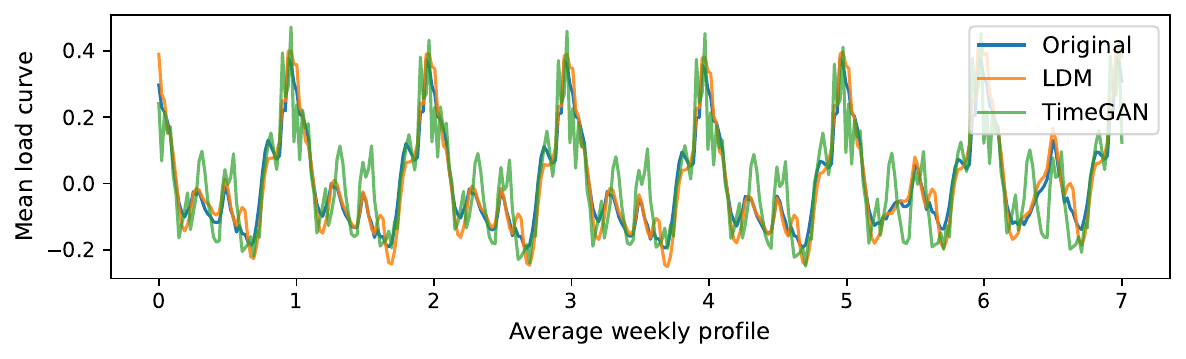}}
\caption{Average weekly profiles of Latent Diffusion (orange) and TimeGAN (green) against test data (blue).
}
\label{fig:week-profiles}
\end{figure}

Then, we compute the average weekly consumption of every sample, resulting in a 336-dimensional vector, and plot these mean profiles per label conditioning.
A representative set of such profiles is shown in Figure \ref{fig:week-profiles}.
Different time-of-use rates induce strong weekly patterns, which are accurately reproduced by latent diffusion for the night and midday rates.
This illustrates the flexibility of the conditioning mechanism, since label information is only included in the diffusion model and not in learning the autoencoder.
On the contrary, TimeGAN's profiles are noisier, although they capture the trend.
We note that the \emph{misc.} ToU patterns are harder to reproduce, possibly because they gather more diverse rate plans and because they represent about 10\% of the training and test data.

\begin{figure}[]
\centering
\subfigure[6kVA, night]{\includegraphics[width=0.3\textwidth]{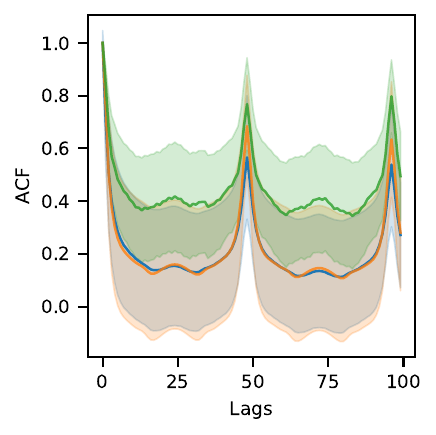}}
\subfigure[6kVA, midday]{\includegraphics[width=0.3\textwidth]{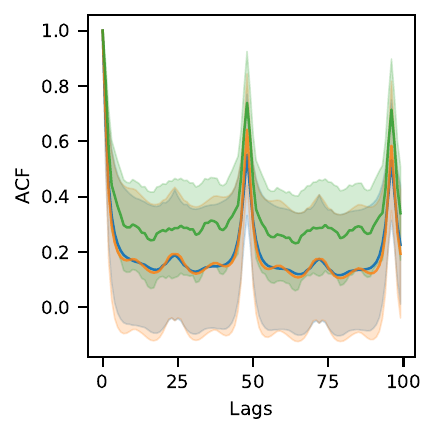}}
\subfigure[9kVA, night]{\includegraphics[width=0.3\textwidth]{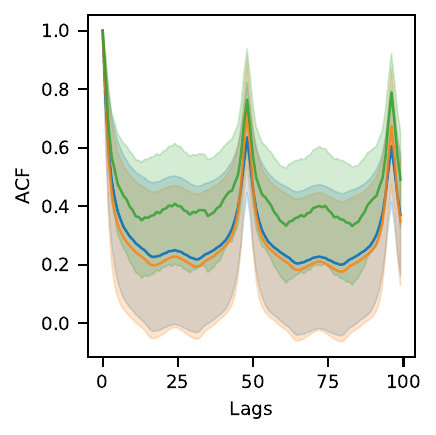}}
\subfigure[9kVA, misc.]{\includegraphics[width=0.3\textwidth]{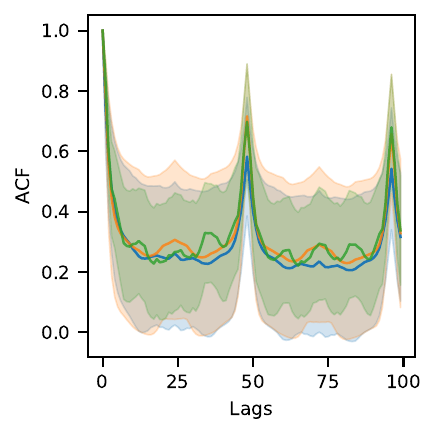}}
\subfigure[12kVA, midday]{\includegraphics[width=0.3\textwidth]{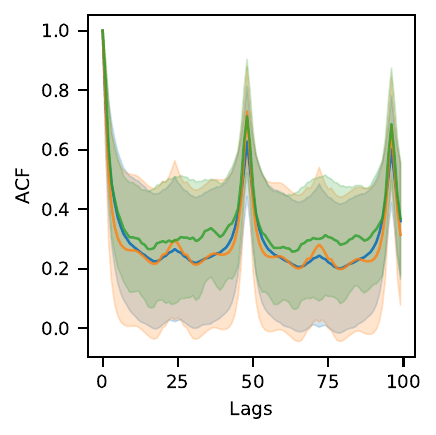}}
\subfigure[All labels]{\includegraphics[width=0.3\textwidth]{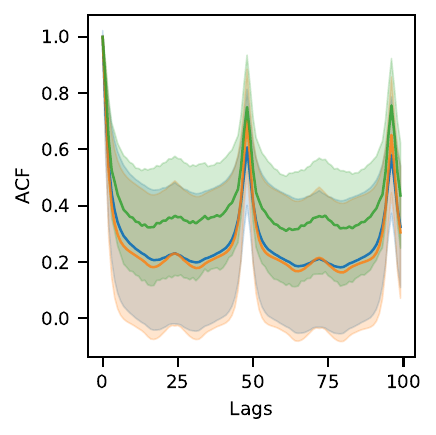}}
\caption{Average autocorrelation functions of Latent Diffusion (orange) and TimeGAN (green) against test data (blue).
Shaded areas denote $\pm$ standard deviation across samples.
}
\label{fig:acf}
\end{figure}

As for temporal consistency, both Latent Diffusion and TimeGAN accurately capture the daily periodicity in electricity consumption (peak at lag 48), which is shown by the plots of the autocorrelation functions in Figure \ref{fig:acf}.
However, TimeGAN adds spurious correlations at other lags, whereas Latent Diffusion shows high fidelity, both in mean and variance.

\subsubsection{Thermo-sensitivity} \label{app:results-thermo}

\begin{figure}
\centering
\subfigure[6kVA, night]{\includegraphics[width=0.275\textwidth]{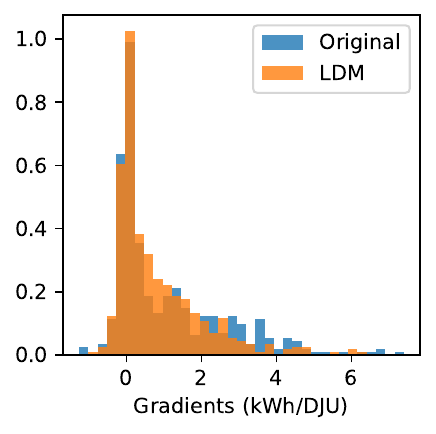}}
\hspace{1ex}
\subfigure[9kVA, midday]{\includegraphics[width=0.275\textwidth]{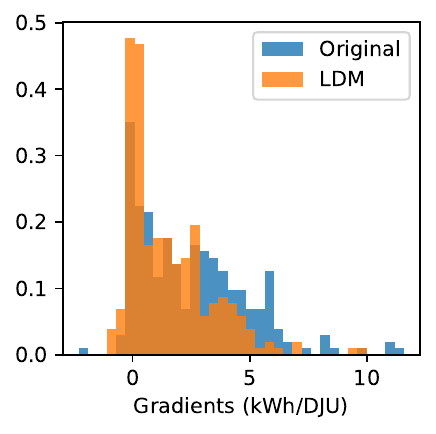}}
\hspace{1ex}
\subfigure[All labels]{\includegraphics[width=0.275\textwidth]{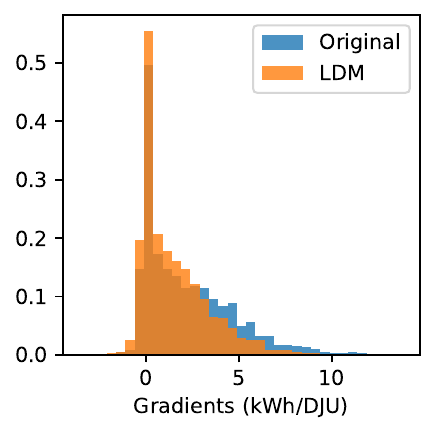}} \\
\subfigure[6kVA, night]{\includegraphics[width=0.275\textwidth]{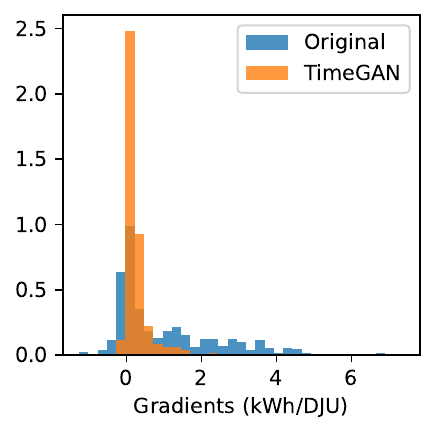}}
\hspace{1ex}
\subfigure[9kVA, midday]{\includegraphics[width=0.275\textwidth]{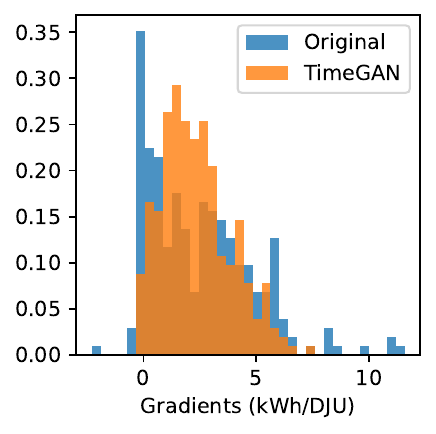}}
\hspace{1ex}
\subfigure[All labels]{\includegraphics[width=0.275\textwidth]{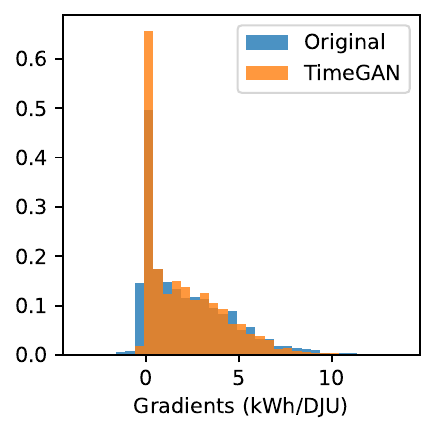}} 
\caption{Thermo-sensitivity gradients computed from load curves aggregated at a daily frequency.
Top: Latent Diffusion, bottom: TimeGAN.
}
\label{fig:thermo-grads}
\end{figure}

\begin{figure}
\centering
\subfigure[]{\includegraphics[width=0.475\textwidth]{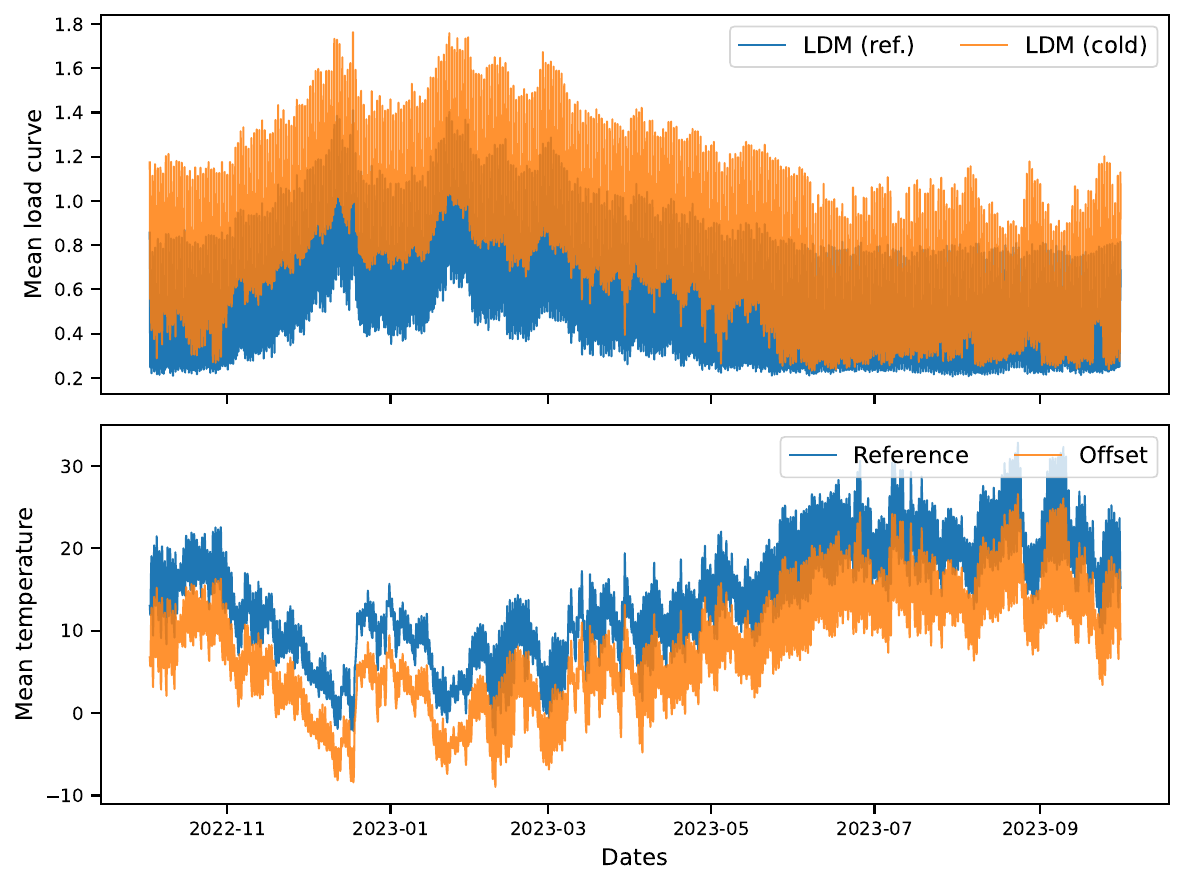}}
\subfigure[]{\includegraphics[width=0.475\textwidth]{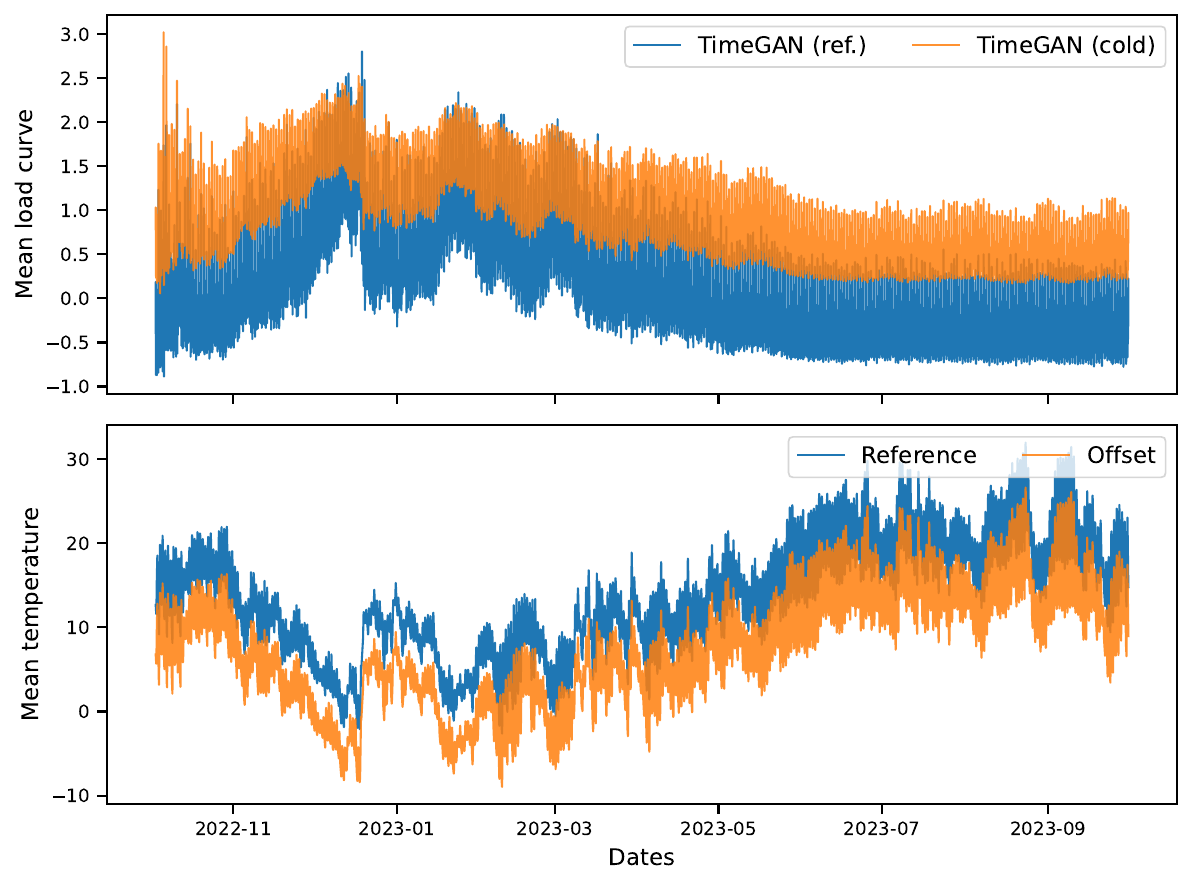}}
\caption{
Average one-year synthetic load curves in the class 6 kVA, night ToU, with (orange) or without (blue) temperature offset for (a) Latent Diffusion and (b) TimeGAN.
}
\label{fig:thermo-offset}
\end{figure}

\paragraph{Gradients}
Figure \ref{fig:thermo-grads} shows some thermo-sensitivity gradients -- see Section \ref{gradient} -- for the two models, compared to the hold-out test distribution.
Load curves generated by Latent Diffusion have realistic gradients, matching real data.
On the contrary, there is a slight mismatch on certain classes, e.g. for the category 6 kVA and night ToU, for TimeGAN.

\paragraph{Temperature offset}
Qualitatively, we assess how the synthetic load curves are distorted, in average, when generating under the same conditioning except that the temperature undergoes a strong \SI{-6.25}{\celsius} offset throughout the year.
We expect the consumption
\begin{enumerate*}[label=(\roman*)]
\item to increase during winter days, since it is mostly driven by heating;
\item to increase in summer, due to the strong offset, but less than in winter.
\end{enumerate*}
Latent Diffusion samples in Figure \ref{fig:thermo-offset} approximately follow this trend, particularly for winter days.
There is a significant increase in summer as well, although the minimum values with or without offset are roughly the sames.
For TimeGAN, the strong temperature offset translates into a strong increase of the minimum electricity consumption all year long.
However, the peak consumption is the same with or without offset during winter, which seems counter-intuitive if not unrealistic.

\subsubsection{Individual samples}

\begin{figure}
\centering
\includegraphics[width=0.95\textwidth]{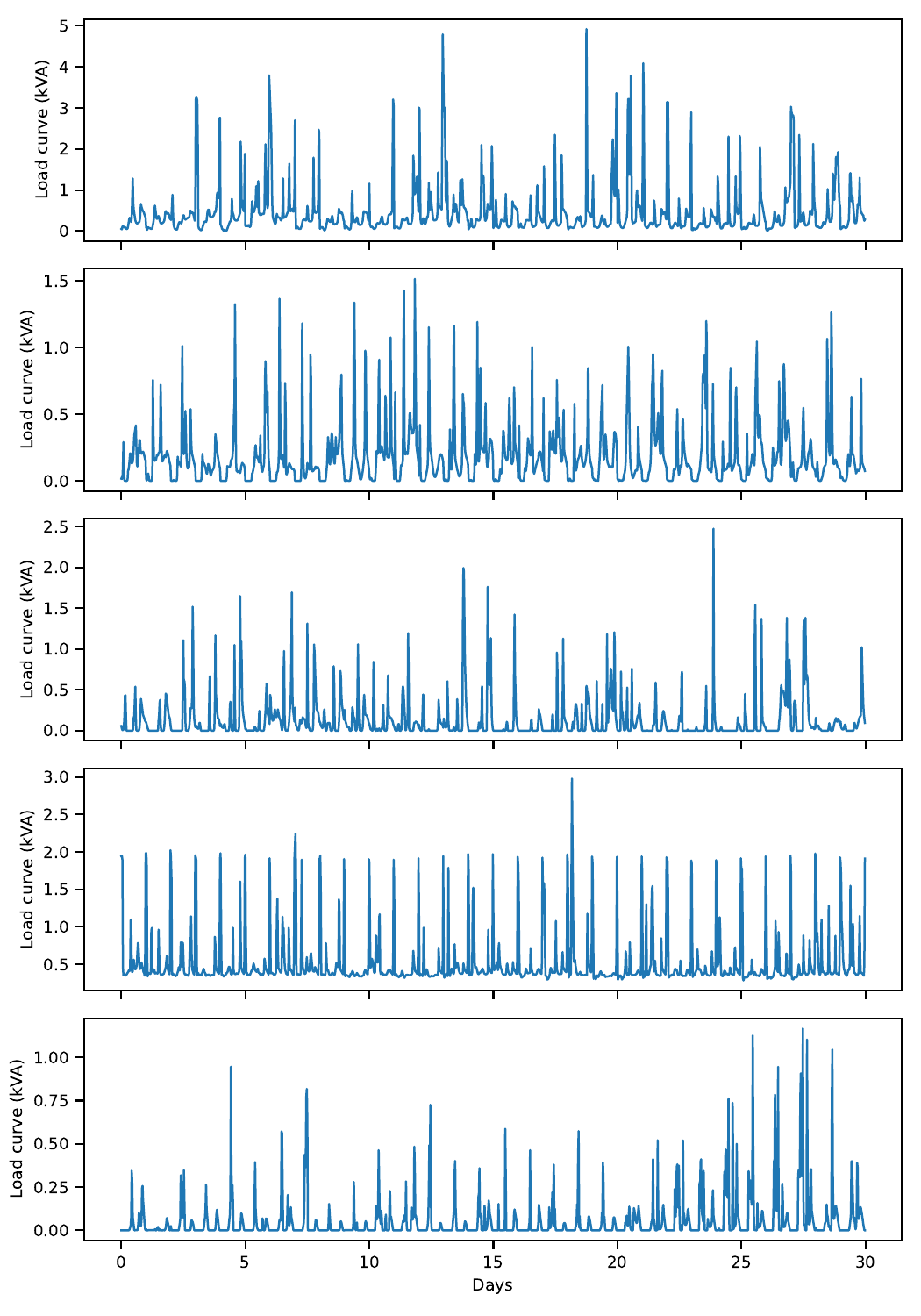}
\caption{Individual samples (contracted power 6 kVA).}
\label{fig:samples6}
\end{figure}

\begin{figure}
\centering
\includegraphics[width=0.95\textwidth]{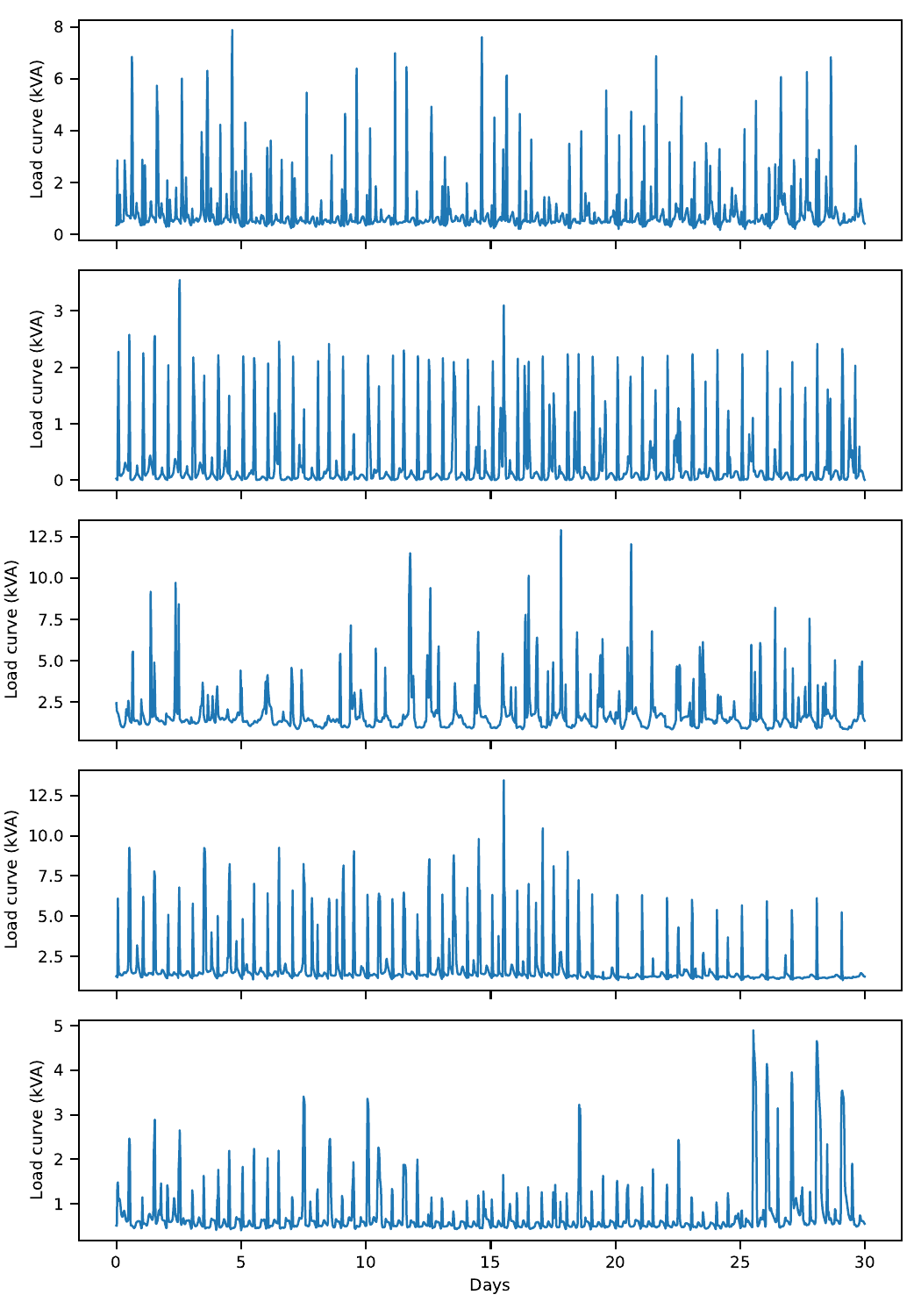}
\caption{Individual samples (contracted power 9 kVA).}
\label{fig:samples9}
\end{figure}

\begin{figure}
\centering
\includegraphics[width=0.95\textwidth]{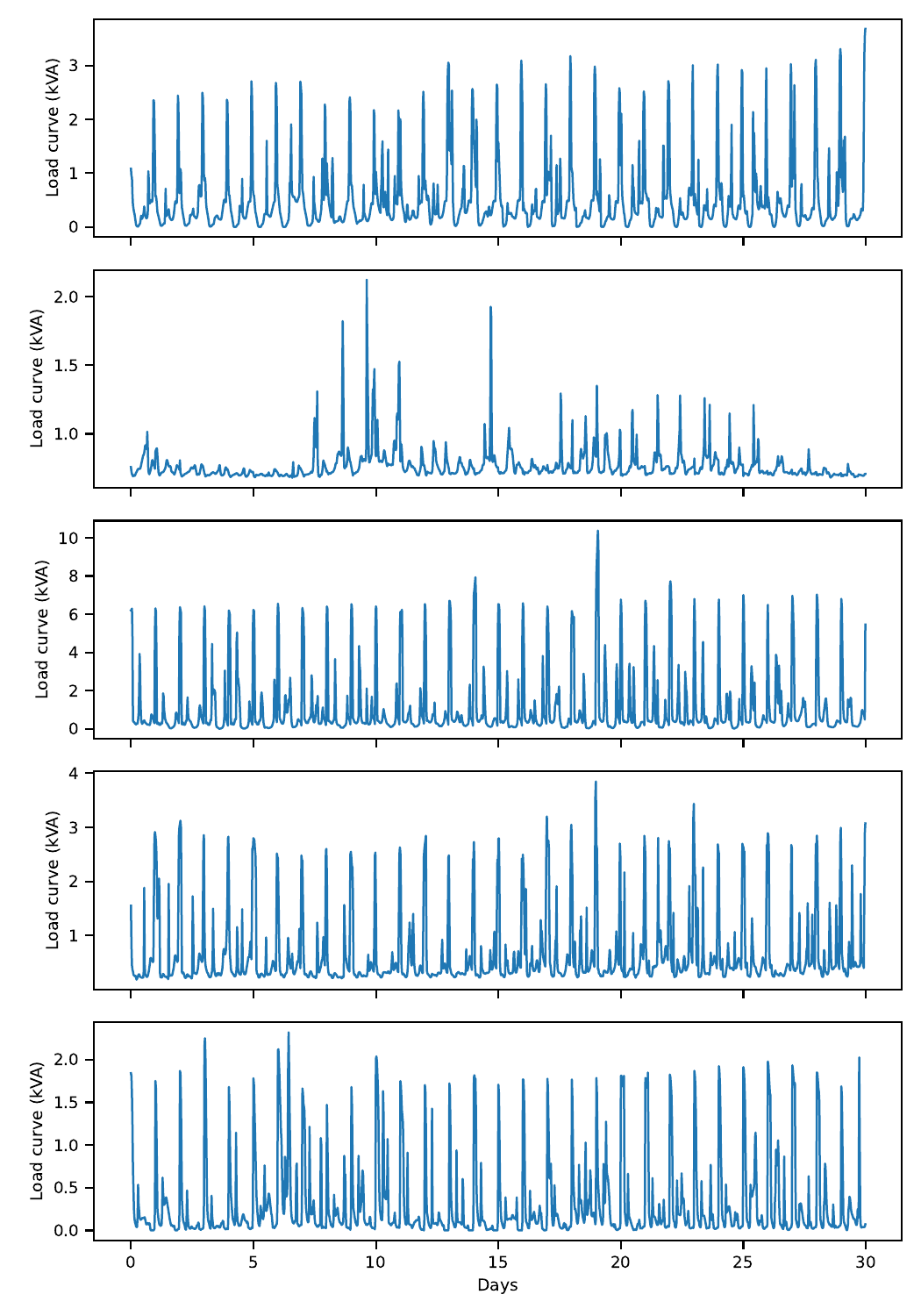}
\caption{Individual samples (contracted power 12 kVA).}
\label{fig:samples12}
\end{figure}

Finally, a random set of load curves generated by latent diffusion are shown in Figures \ref{fig:samples6}, \ref{fig:samples9} and \ref{fig:samples12}.
Although it is challenging for the untrained eye to determine whether a given sample is realistic or not, it can be noted that the samples exhibit clear patterns, s.a. daily peaks, constant periods at low consumption, intermediate plateaus.
The samples also show some diversity.
To the best of our knowledge, these samples make thus viable load curves.


\subsection{Utility}

The full results for evaluating the forecasting utility are shown in Table \ref{tab:forecast-full}.
Load curves having known weekly seasonalities, we implement a simple baseline for the sake of comprehensiveness.
To predict the next $H$ values, the baseline repeats the last $H$ values from one week ago, i.e. $\hat{\mathbf{x}}_{t+1:t+H}=\mathbf{x}_{t-336+1:t-336+H}$, for $H\leq336$.
Predicting individual load curves at sub-hourly frequency is notoriously challenging, particularly with no exogenous variables, due to the stochastic nature of user behaviors.
This can be seen from the relatively modest improvement in mean absolute error (MAE) of PatchTST trained on Latent Diffusion over the baseline.
Nevertheless, on both MSE and MAE and across all horizons, PatchTST trained on Latent Diffusion samples
\begin{enumerate*}[label=(\roman*)]
\item is quasi-equivalent to its counterpart on real samples and
\item outperforms the model trained on TimeGAN.
\end{enumerate*}

\begin{table}
\centering
\caption{Forecasting from a lookback of 720.
Closest to TRTR (Train on Real, Test on Real) is emphasized in bold.
}
\begin{tabular}{cc|ccc|c}
\toprule
\bfseries Horizon & \bfseries Loss & \bfseries TRTR & \bfseries LDM & \bfseries TimeGAN & \bfseries Repeat \\
%\midrule
%$H$ & \emph{Loss} & \multicolumn{5}{c}{\emph{Forecasting}} \\
\midrule
\multirow{2}{*}{48} & \small MSE & 0.204  & \bfseries 0.204 &0.225 & 0.334 \\
 & \small MAE & 0.240 &  \bfseries 0.237 & 0.259 & 0.264 \\
\midrule
\multirow{2}{*}{96} & \small MSE & 0.188  & \bfseries0.188 & 0.207 & 0.308 \\
 & \small MAE & 0.230 &  \bfseries 0.228 & 0.250 & 0.252 \\
\midrule
\multirow{2}{*}{192} & \small MSE & 0.177 & \bfseries 0.177 & 0.193 & 0.290 \\
 & \small MAE & 0.228 &  \bfseries 0.227 & 0.245 & 0.243 \\
\midrule
\multirow{2}{*}{336} & \small MSE & 0.192 & \bfseries 0.192 & 0.211 & 0.294 \\
 & \small MAE & 0.239 &  \bfseries 0.239 & 0.257 & 0.246\\
\bottomrule
\end{tabular}
\label{tab:forecast-full}
\end{table}

\subsection{Privacy} \label{app:res-privacy}

\begin{figure}[b]
\centering
\subfigure[Black-box]{\includegraphics[width=0.35\textwidth]{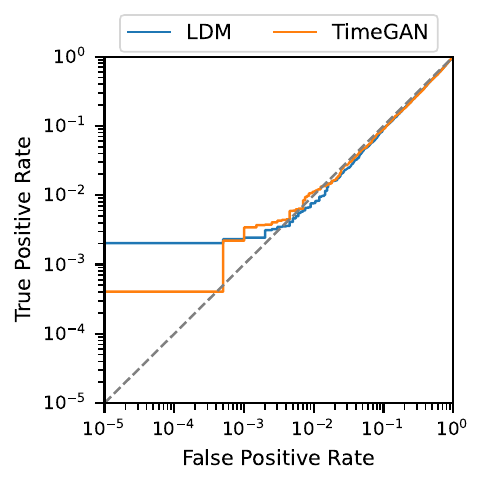}}
\hspace{2ex}
\subfigure[White-box]{\includegraphics[width=0.35\textwidth]{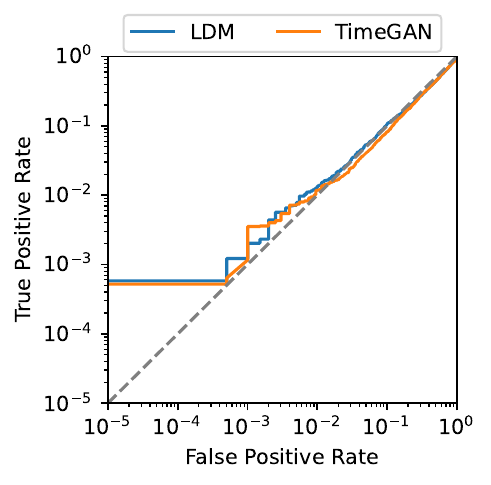}}
\caption{ROC curves in log scale of the membership inference attacks in the (a) black-box and (b) white-box scenarios.}
\label{fig:mia-roc}
\end{figure}

\paragraph{MIA}
We plot the ROC curves in Figure \ref{fig:mia-roc} and report the black-box and white-box scores, namely the true positive ratio at 0.1\% false positive ratio for membership inference attacks in Table \ref{tab:mia-privacy}, to summarize them  \citep{carlini2022membership}.
Latent diffusion and TimeGAN obtain similar metrics:
all scores are close to 0, typically around 0.002 (0.2\%) for black-box and 0.1\% for white-box, indicating a limited risk of re-identification.
For the black-box attack, we also investigate whether the size of the available synthetic set, from which the smallest distance is computed, affects the scores.
Except for very small sizes, Table \ref{tab:mia-privacy} shows little variations in the scores, with a slight downward trend.
Overall, the attacks on both LDM and TimeGAN are close to random and fail to reliably identify a subset of users.

\begin{table}
\centering
\caption{True Positive Ratio ($\times 10^3$) at 0.1\% False Positive Ratio for the membership inference attacks.
The scores of the black-box attacks are averaged over 5 random subsamples (with $\pm$ standard deviation) of size $n_{gen}$ of synthetic samples.
}
\resizebox{\textwidth}{!}
{
\begin{tabular}{r|ccccc|c}
\toprule
& \multicolumn{5}{c|}{\bfseries Black-box} & \bfseries White-box \\
\midrule
$n_{gen}$ & 10 & 500 & 1k & 1.5k & 2k & - \\
\midrule
\bfseries LDM     & 0.43\scriptsize$(\pm0.36)$ & 2.66\scriptsize$(\pm0.50)$ & 2.64\scriptsize$(\pm0.37)$ & 2.47\scriptsize$(\pm0.11)$ & 2.26\scriptsize$(\pm0.13)$ & 1.216 \\
\bfseries TimeGAN & 1.19\scriptsize$(\pm0.74)$ & 2.59\scriptsize$(\pm0.36)$ & 2.52\scriptsize$(\pm0.11)$ & 2.31\scriptsize$(\pm0.10)$ & 2.21\scriptsize$(\pm0.03)$ & 0.637 \\
\bottomrule
\end{tabular}
}
\label{tab:mia-privacy}
\end{table}

\paragraph{Three-sample MMD Test} \label{app:mmd}
The null hypothesis $\mathcal{H}_0$ is that the synthetic set of load curves is closer to the real test set than to the real train set.
We compute the p-values of the three-sample MMD-test for all 9 categories of time-of-use $\times$ contracted power, as well as for the entire synthetic dataset, in Table \ref{tab:mmd-privacy}.
The p-values are reasonably large for all categories and both Latent Diffusion and TimeGAN, suggesting not to reject $\mathcal{H}_0$.
Hence, this test does not give evidence that the models overfit the training data.

\begin{table}[]
\centering
\caption{p-values for the MMD three-sample test \citep{bounliphone2016test}.}
\resizebox{\textwidth}{!}
{
\begin{tabular}{r|ccc|ccc|ccc|c}
\toprule
\bfseries ToU & \multicolumn{3}{c|}{Midday} & \multicolumn{3}{c|}{Night} & \multicolumn{3}{c|}{Misc} & all\\
\midrule
\bfseries Power (kVA)& 6 & 9 & 12 & 6 & 9 & 12 & 6 & 9 & 12 & all\\ 
\midrule
\bfseries LDM     & 0.297 & 0.930 & 0.852 & 0.019   & 0.218 & 0.795 & 0.239 & 0.325 & 0.642 & 0.845 \\
\bfseries TimeGAN & 0.304 & 0.953 & 0.996 & 0.952 & 0.813 & 0.840 & 0.043 & 0.341 & 0.366 & 0.948 \\
\bottomrule
\end{tabular}
}
\label{tab:mmd-privacy}
\end{table}

\begin{figure}
\centering
\subfigure[LDM, year]{\includegraphics[width=0.23\textwidth]{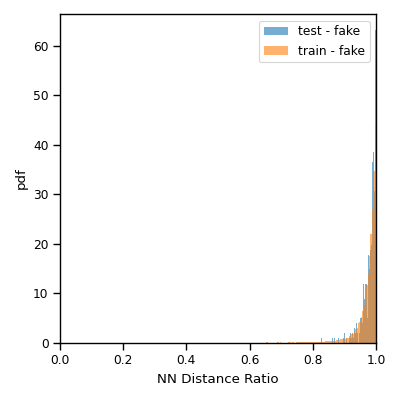}}
\subfigure[LDM, day]{\includegraphics[width=0.23\textwidth]{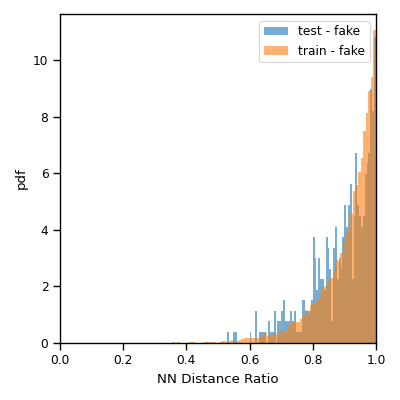}}
\subfigure[TimeGAN, year]{\includegraphics[width=0.23\textwidth]{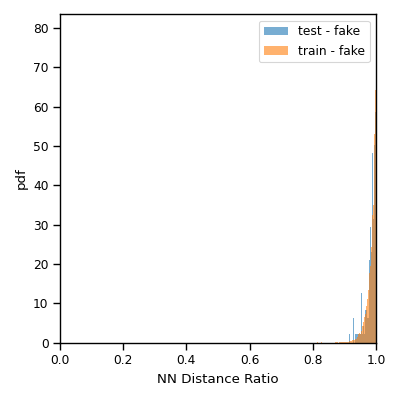}}
\subfigure[TimeGAN, day]{\includegraphics[width=0.23\textwidth]{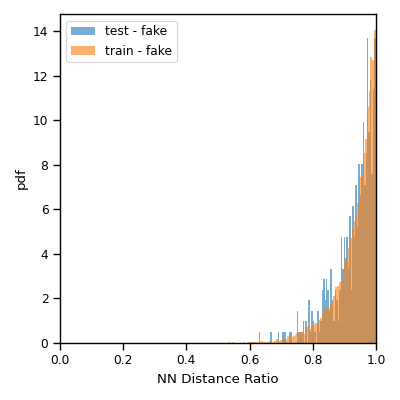}}
\caption{Nearest neighbor distance ratio for Latent Diffusion and TimeGAN, computed either on one-year data or average daily profiles.}
\label{fig:nndr}
\end{figure}

\paragraph{Nearest Neighbor Distance Ratio} 
For both Latent Diffusion and TimeGAN, the distribution of NNDR computed with either real train or test data is the same, as evidenced in Figure \ref{fig:nndr}.
They are skewed towards 1, suggesting that the synthetic samples are not located towards outliers in the training set, which would constitute a breach of privacy.

\end{document}